\documentclass[runningheads]{llncs}
\usepackage{graphicx}

\usepackage{tikz}
\usepackage{comment}
\usepackage{amsmath,amssymb} 
\usepackage{mathtools}
\usepackage{multirow}
\usepackage{mathabx}
\usepackage{color}
\usepackage{cite}
\usepackage{url}
\usepackage{microtype}
\usepackage{bbm}

\usepackage[accsupp]{axessibility}  

\definecolor{bananayellow}{rgb}{1.0, 0.88, 0.21}
\definecolor{chromeyellow}{rgb}{1.0, 0.65, 0.0}

\begin{document}
\pagestyle{headings}
\mainmatter
\def\ECCVSubNumber{3512}  

\title{D2HNet: Joint Denoising and Deblurring with Hierarchical Network for Robust Night Image Restoration} 

\titlerunning{D2HNet}
%
\author{Yuzhi Zhao\inst{1}
\and
Yongzhe Xu\inst{2}
\and
Qiong Yan\inst{2}
\and
Dingdong Yang\inst{2}
\and
Xuehui Wang\inst{3}
\and
Lai-Man Po\inst{1}
}
\authorrunning{Y. Zhao et al.}
%
\institute{Department of Electrical Engineering, City University of Hong Kong
\and
SenseTime Research and Tetras.AI
\and
MoE Key Lab of Artificial Intelligence, AI Institute, Shanghai Jiao Tong University \\
\email{yzzhao2-c@my.cityu.edu.hk, wangxuehui@sjtu.edu.cn, eelmpo@cityu.edu.hk, \{xuyongzhe1,yanqiong,yangdingdong\}@tetras.ai}
}
\maketitle

\begin{abstract}

Night imaging with modern smartphone cameras is troublesome due to low photon count and unavoidable noise in the imaging system. Directly adjusting exposure time and ISO ratings cannot obtain sharp and noise-free images at the same time in low-light conditions. Though many methods have been proposed to enhance noisy or blurry night images, their performances on real-world night photos are still unsatisfactory due to two main reasons: 1) Limited information in a single image and 2) Domain gap between synthetic training images and real-world photos (e.g., differences in blur area and resolution). To exploit the information from successive long- and short-exposure images, we propose a learning-based pipeline to fuse them. A D2HNet framework is developed to recover a high-quality image by deblurring and enhancing a long-exposure image under the guidance of a short-exposure image. To shrink the domain gap, we leverage a two-phase DeblurNet-EnhanceNet architecture, which performs accurate blur removal on a fixed low resolution so that it is able to handle large ranges of blur in different resolution inputs. In addition, we synthesize a D2-Dataset from HD videos and experiment on it. The results on the validation set and real photos demonstrate our methods achieve better visual quality and state-of-the-art quantitative scores. The D2HNet codes and D2-Dataset can be found at \url{https://github.com/zhaoyuzhi/D2HNet}.


\end{abstract}

\section{Introduction}

Capturing high-quality photos at night-time on modern smartphones is troublesome due to the limitations of sensors and optical systems. It is a long-standing and practical problem in the computational photography field. Acquiring sharp and clean photos effectively and efficiently on smartphones in night conditions is in great demand. The main difficulty lies in that the image signal is too weak compared with the inherent noise in the imaging process, which yields a low signal-to-noise ratio (SNR) and degrades image quality \cite{li2018structure, wang2019progressive, liba2019handheld}. To obtain higher SNR, there are many solutions either on the hardware level (in-camera solutions) or algorithm level, which typically fall into one of these three categories: 1) \emph{Physical solutions}: using a larger sensor, opening the aperture, using flash, or setting longer exposure time; 2) \emph{Single-image restoration}: deblurring the long-exposure image with motion blurs, or denoising the short-exposure image with severe noises; 3) \emph{Burst-image restoration}: combining several photos captured in quick succession using temporal coherence within the burst.

Though these solutions improve the night image restoration quality, they might not meet the requirements of both \emph{effectiveness} and \emph{efficiency} for mobile photography. For \emph{physical solutions}, larger sensor size and aperture are related to hardware design and increase the cost. The built-in flash does not help for far scenes. Long exposure time causes motion blur. To post-process the captured images, \emph{single-image restoration} methods have been widely studied, e.g., training neural networks \cite{kupyn2018deblurgan, zhang2017beyond} on a large number of paired degraded-clean images. However, a single input image contains limited information thus restricting the restoration quality. To use more information, \emph{burst-image restoration} methods \cite{liu2014fast, mildenhall2018burst} combine multiple continuous frames to generate a single good image. Though they have a theoretically superior SNR than single-image restoration methods, the speed is restricted by the capturing process, including multiple exposure and readout time. Meanwhile, the misalignment issue has to be solved for all captured frames. In addition, the data distribution gap (e.g., differences in blur area, resolution) between training and real images remains a key problem.

In this paper, we tackle this real-world problem by post-processing successive long- and short-exposure images through a D2HNet framework. It can produce clean and sharp photographs on mobile devices without any manual control or extra hardware support. Compared with the previous image restoration approaches, our approach has three main advantages: 1) Taking advantage of both long and short exposures; 2) Addressing the domain gap issue between training data and real-world photos by a special two-phase network; 3) Balancing image processing quality and capturing time (only 2 long- and short-exposure images are needed).

\begin{figure}[t]
\centering
\includegraphics[width=118mm, height=56mm]{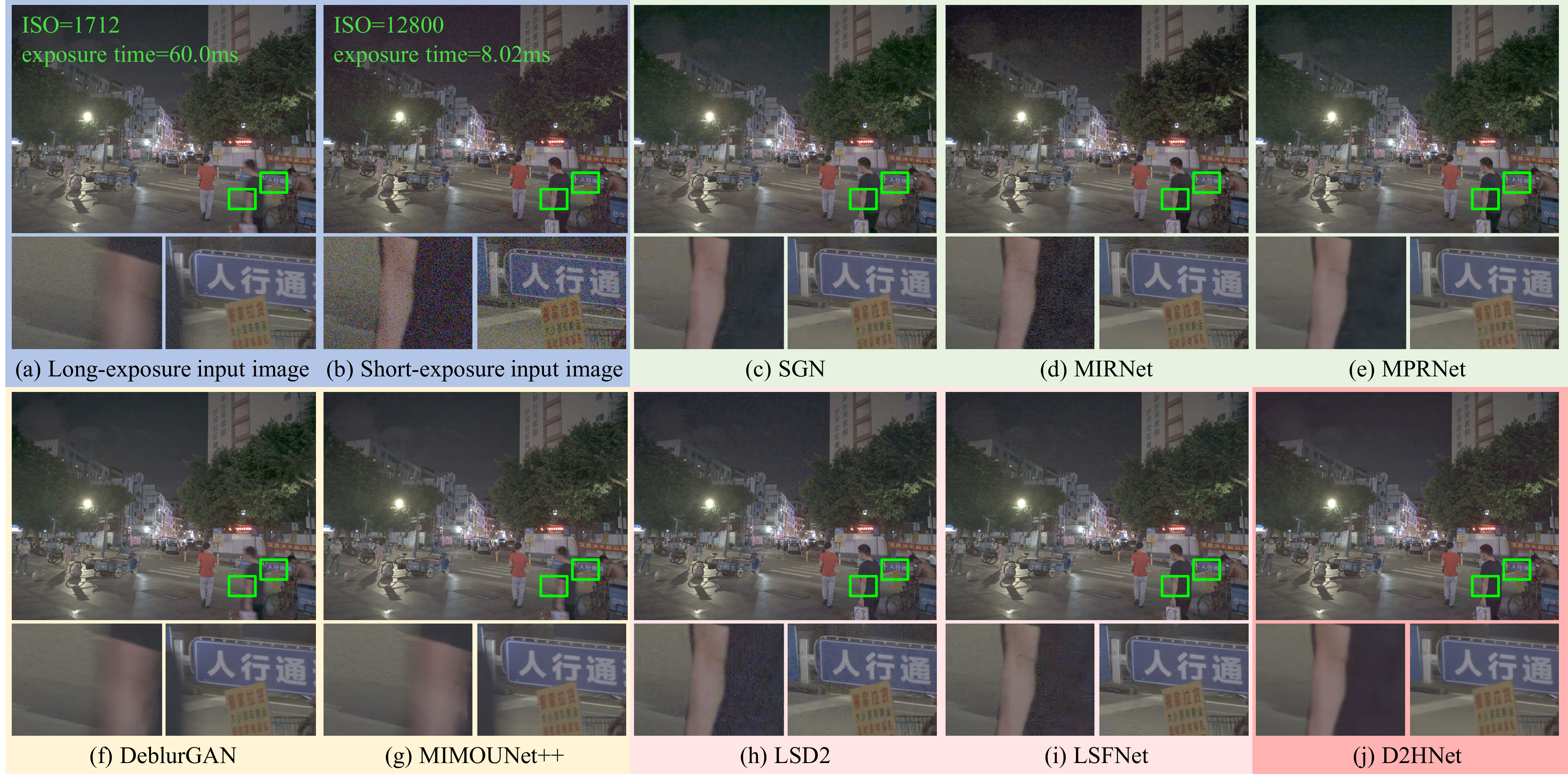}

\caption{Performance on real-world night photos. Input photos captured by \emph{Xiaomi Mi Note 10} smartphone are shown in (a) and (b). Results from single image denoising methods are in (c) - (e). Results from single image deblurring methods are in (f) and (g). Results from long-short fusion methods (including our D2HNet) are in (h) - (j).}
\label{teaser}

\end{figure}

Normally, long-exposure images have regular color and fewer noises and short-exposure images are of trivial blurs. Compared with single-image restoration \cite{kupyn2018deblurgan, zhang2017beyond}, fusing them helps reduce the noise level and blurriness, and improve color fidelity for night photos. Compared with burst-image restoration methods \cite{liu2014fast, mildenhall2018burst}, two shots have milder misalignment issues and require a shorter capturing time. Built upon these observations, we propose the D2HNet framework. To address the domain gap between training data and real photos captured by different smartphones, we split D2HNet into sequential subnets: \emph{DeblurNet} and \emph{EnhanceNet}. DeblurNet runs on a small resolution at the training and a fixed resolution at the testing similar to training images. It ensures pixel shifts or blur levels between training and real-world images are comparable; therefore, the network generalizes better to different image resolutions and blur areas. EnhanceNet enhances the DeblurNet output on actual target resolutions together with short- and long-exposure inputs. We use deformable convolutions \cite{dai2017deformable} in the EnhanceNet to align the features hierarchically to better extract the details and textures from the long-exposure input. In addition, we propose a \emph{CutNoise scheme} to assist the learning of where and how to deblur and a \emph{VarmapSelection scheme} to balance blurry and non-blurry patches during training.

To evaluate the capability of D2HNet, we synthesize a D2-Dataset from HD videos for training and validation. It contains 6853 tuples of long- and short-exposure images with corresponding sharp ground truth. It covers a wide range of scenes, e.g., cities, villages, forests, deserts, and mountains. We also capture 28 pairs of long- and short-exposure photos of real-world scenes by a smartphone for testing. Extensive experiments on both D2-Dataset and real captured photos show the state-of-the-art (SOTA) performance achieved by the proposed D2HNet. One real sample is shown in Figure \ref{teaser}. D2HNet can produce clean and sharp images simultaneously, while the other methods fail to do so. It demonstrates that D2HNet better utilizes the information of dual-exposure images.

Below we summarize the main contributions of this paper:

1) We propose a two-phase D2HNet for robust real night image restoration and to address the domain gap issue between training data and real photos;

2) We propose two data augmentation schemes, CutNoise and VarmapSelection, to improve and stabilize the training of D2HNet;

3) We create a D2-Dataset including 6853 image tuples with multiple levels of blurs for benchmarking D2HNet;

4) We conduct extensive experiments with long-short fusion methods, and single image denoising or deblurring methods. The proposed D2HNet achieves better performance than other methods.

\section{Related Work}

\textbf{Single-image Denoising.} Image denoising is a fundamental topic in image processing. Previous methods such as total variation \cite{rudin1992nonlinear}, wavelet coring \cite{simoncelli1996noise}, non-local means \cite{buades2005non}, BM3D \cite{dabov2007image} assumed noises and signals have specific statistical regularities. However, these methods used hand-crafted models thus not robust to real noises. Recently, CNNs have shown their advanced performance to address blind denoising issue \cite{mao2016image, zhang2017beyond, tai2017memnet, chen2016trainable, zhang2018ffdnet, liu2018multi, gu2019self, liu2020densely}. Some works further extended them to reduce real noises \cite{zhang2017fast, chen2018image, chen2018learning, anwar2019real, guo2019toward, yue2019variational, zamir2020learning, kim2020transfer, zhang2020residual, liu2021invertible, byun2021fbi, hu2021pseudo, cheng2021nbnet, zamir2021multi, lamba2021restoring, chen2021hinet}. To better simulate noise emerged on mobile ISP, many inverse algorithms \cite{brooks2019unprocessing, zamir2020cycleisp, xing2021invertible} and real noise calibration methods \cite{abdelhamed2019noise, wei2020physics, wang2020practical} were proposed.


\noindent \textbf{Single-image Deblurring.} Image deblurring aims to generate a sharp and clean reconstruction from a blurry input. Many classical non-blind methods formulated the problem as blind deconvolutions \cite{richardson1972bayesian, krishnan2009fast, levin2011efficient}. The blur kernels are normally assumed as noisy linear operators enforced on the clean images. Recently, CNN-based approaches \cite{sun2015learning, chakrabarti2016neural, gong2017motion, nimisha2017blur, nah2017deep, zhang2018dynamic, tao2018scale, kupyn2018deblurgan, kupyn2019deblurgan, gao2019dynamic, purohit2020region, suin2020spatially, rim2020real, park2020multi, yuan2020efficient, zhang2021exposure, cho2021rethinking, ji2022xydeblur, whang2022deblurring} proposed the end-to-end deblurring with specific network architectures and loss functions. These methods are trained on large-scale blurry-sharp pairs. However, directly applying them to real-world photos may not obtain sharp results.

\noindent \textbf{Burst-image Restoration.} Since the overall photon counts of burst images are more than a single image, burst-image-based methods \cite{liu2014fast, mildenhall2018burst, godard2018deep, xu2019learning, xia2020basis, zhang2020attention, karadeniz2021burst, dudhane2022burst} have theoretically superior SNR than single-image-based methods. However, burst images suffered from noises and camera shake, which increase the difficulty of implementation. To overcome that, \cite{godard2018deep} proposed a recurrent neural network to filter noises in a sequence of images. \cite{mildenhall2018burst} combined neural network and kernel method to perform denoising and alignment jointly. Though they restore high-quality photos, their data capture occupies a major time during application.

\noindent \textbf{Image Restoration by Fusing Successive Long- and Short-exposure images.} Image restoration with dual exposures \cite{yuan2007image, choi2008motion, tico2010motion, son2011pair, whyte2012non, son2013image, gu2020blur} is beneficial for both noise reduction and blur estimation. For instance, Yuan et al. \cite{yuan2007image} firstly estimated blur kernels using the texture of short-exposure images, which are then used to restore the long-exposure blurry images. Recently, LSD2 \cite{mustaniemi2020lsd} and LSFNet \cite{chang2021low} used CNNs to fuse dual-exposure images and obtained better results than single-image denoising or deblurring methods on their synthetic dataset. However, they ignored the potential domain gap issue between training images and real-world photos.

\noindent \textbf{Deformable Convolution.} Dai et al. \cite{dai2017deformable} proposed deformable convolutions, which allows the network to obtain the information away from regular local neighborhoods by learning additional offsets. It has been widely applied in computer vision tasks such as semantic segmentation \cite{dai2017deformable, zhu2019deformable}, video deblurring \cite{wang2019edvr}, video super-resolution \cite{tian2020tdan, chan2021basicvsr, chan2022basicvsr++}, and video restoration \cite{deng2020spatio, guo2022differentiable}. For instance, EDVR \cite{wang2019edvr} used deformable convolutions to align inputs without using explicit optical flows. For the long-short fusion problem, there normally exists a misalignment issue between input long- and short-exposure images. Also, it is difficult to compute accurate optical flows from noisy and blurry inputs. Inspired by previous methods, we adopt deformable convolutions as alignment blocks.

\section{Data Acquisition}

\textbf{D2-Dataset.} We synthesize a D2-Dataset for training and benchmarking. The data synthesis pipeline is as follows:

1) We collect 30 HD videos with 1440$\times$2560 resolution from the Internet. They are almost noise-free and cover a wide range of scenes. We sample 60 continuous frames (approximately 1 second in original 60-fps videos) every 10 seconds in each video to reduce repeated scenes and avoid scene switching;

2) We use a video frame interpolation model SuperSloMo \cite{jiang2018super} to increase the original 60-fps videos to 960 fps. It smooths videos to simulate realistic blurs;

3) We synthesize successive long-exposure image $l$ and short-exposure image $s$ by averaging interpolated frames. Meantime, we add a time gap between $l$ and $s$ to model hardware readout limitation. We also extract corresponding sharp single frames, i.e., the last frame of long-exposure image $l_{last}$ and the first frame of short-exposure image $s_{first}$.

The pipeline results in 6853 image tuples ($l$, $s$, $l_{last}$, and $s_{first}$), where 5661 tuples are used for training and 1192 for validation. More details are presented in the supplementary material.

\noindent \textbf{Testing Images.} We capture 28 pairs of long- and short-exposure images with resolution 3472$\times$4624 using a \emph{Xiaomi Mi Note 10 smartphone}. To ensure the overall brightness of long- and short-exposure images are approximately the same, we set ``ISO$\times$exposure time'' of them equal. Specifically, the exposure time of the long-exposure image is set to be 8 times of short-exposure time while its ISO is $1/8$ of the short-exposure image.


\section{Methodology}

\subsection{Problem Formulation}

Given paired noisy long- and short-exposure images denoted as $l_n$ and $s_n$, we aim to recover a sharp and clean image $z$. We formulate it as maximizing a posteriori of the output conditioned on inputs and D2HNet parameters $\Theta$:
\begin{equation}
\Theta^* = \mathop{\arg\max}\limits_{\Theta} p( z | l_n, s_n, \Theta ).
\label{pf}
\end{equation}

We train our network on the proposed D2-Dataset. From it we use $l$, $s$ pair to generate noisy training inputs $l_n, s_n$, and $s_{first}$ as ground truth of $z$ here.

\subsection{D2HNet Architecture and Optimization}

\begin{figure*}[t]
\centering
\includegraphics[width=\linewidth]{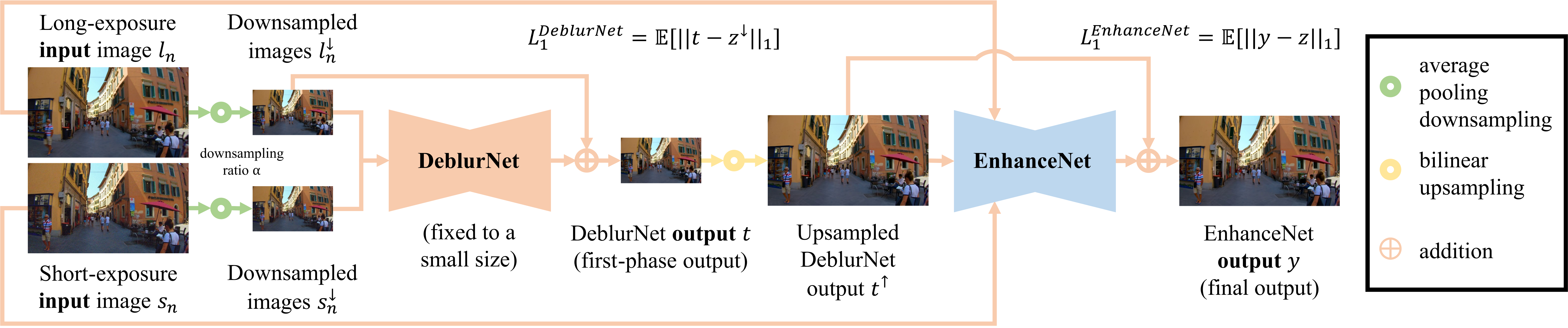}

\caption{Illustration of the D2HNet workflow.}
\label{net1}

\end{figure*}

\begin{figure*}[t]
\centering
\includegraphics[width=\linewidth]{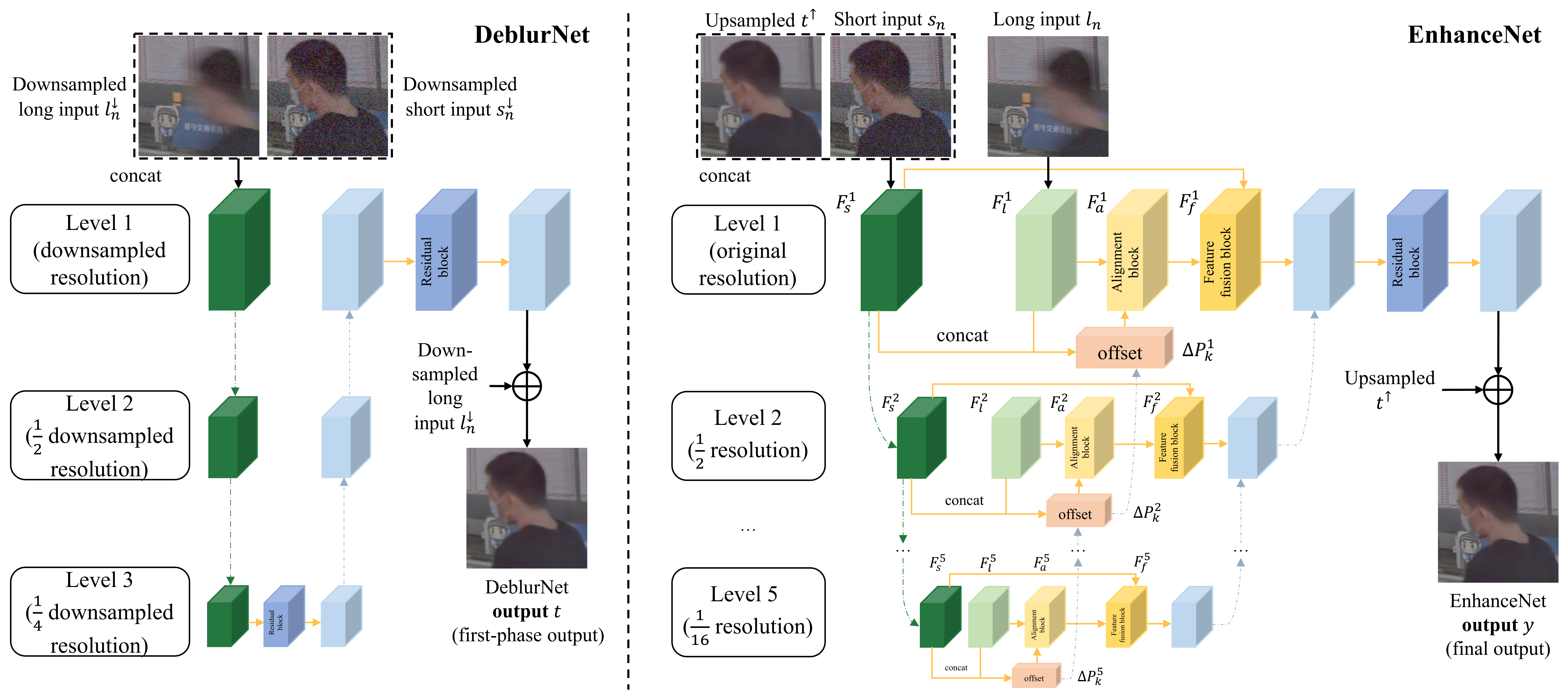}

\caption{Illustration of DeblurNet (left) and EnhanceNet (right). Alignment blocks, Feature fusion blocks, and Residual blocks are noted in the figure while the remaining blocks are normal convolutional layers. Offsets are learnable and as parts of Alignment blocks. Images $t$ and $y$ are the outputs of DeblurNet and EnhanceNet, respectively.}
\label{net2}

\end{figure*}

\textbf{Workflow.} The workflow of D2HNet is illustrated in Figure \ref{net1}. To address the domain gap issue between synthetic training images and real-world photos (e.g., different resolutions and blur levels), we use a two-phase structure in our workflow. In phase one, the two inputs are downsampled into a smaller and fixed resolution, so that motion scales and pixel shifts in the long-exposure image are restricted. Then, the DeblurNet with a certain receptive field can perform accurate deblurring based on the edge information in the short-exposure image. In phase two, to enhance the details lost during downsampling, the EnhanceNet post-process the upsampled first-phase network output together with long-short inputs in their original resolution.

\noindent \textbf{DeblurNet.} For the input data, DeblurNet receives the downsampled long- and short-exposure input images with a downsampling ratio $\alpha$. At the training, $\alpha$ is set to $1/2$; while at the testing, the input resolution is fixed to 1024$\times$1024. Therefore, the domain gap is reduced since the motion scales of testing images are controlled. We use the average pooling as the downsampling operator at both training and testing stages as it mimics the physical differences between low- and high-resolution images better. For the network architecture, DeblurNet uses 3 levels to extract features since it better balances the deblurring quality and computational complexity, where DWT \cite{liu2018multi} is used as the downsampling operator. There are two Residual blocks at the bottleneck and at the tail respectively, where each block includes 4 sequential residual layers \cite{he2016deep}. The output $t$ is upsampled by bilinear sampling and then as the input for the next phase, i.e., EnhanceNet.

\noindent \textbf{EnhanceNet.} The target of EnhanceNet is to recover the details (mostly from long-exposure input) and further remove artifacts for the upsampled DeblurNet output $t^{\uparrow}$. As shown in Figure \ref{net2}, EnhanceNet has 3 modules: feature pyramid extraction (green blocks), alignment and feature fusion (pink and yellow blocks), and reconstruction (blue blocks).

The feature pyramid extraction has two branches without sharing weights. The output two feature pyramids have 5 levels, denoted as $F_s^1$-$F_s^5$ and $F_l^1$-$F_l^5$, respectively. Since inputs $s_n$/$y^{\uparrow}$ and $l_n$ are not spatially aligned, we perform the alignment for long-exposure features ($F_l^1$-$F_l^5$) by Alignment blocks, where we use the modulated deformable convolution \cite{zhu2019deformable}. Alignment blocks allow the following layers to better fuse the information of two feature pyramids. Here we give a brief introduction for the modulated deformable convolution. As we have known, a 3$\times$3 convolution kernel of dilation 1 has learnable weights $w_k \in \{1,...,K\}$ and fixed offsets $p_k \in \{(-1,-1),(-1,0),(-1,1),(0,-1),(0,0),(0,1),(1,-1),(1,0),(1,1)\}$, where $K=9$. Then for the modulated deformable convolution, there are learnable parameters, offsets $\Delta p_k^i$ and modulation scalars $\Delta m_k^i$ for each location $p_k$. The offsets $\Delta p_k^i$ are real numbers and the modulation scalars $\Delta m_k^i$ are in range of [0, 1]. Therefore, for such convolution result on $i$-th long-exposure feature $F_{l}^i$ can be expressed as:
\begin{equation}
F_{a}^i(p) = \sum_{k=1}^{K} w_k^i \cdot F_{l}^i(p + p_k + \Delta p_k^i) \cdot \Delta m_k^i.
\label{deformable}
\end{equation}

The modulation scalars and learnable offsets (pink blocks in Figure \ref{net2}) are learned from short- and long-exposure features hierarchically. For simplicity, we only express the learnable offsets from a series of convolutional layers $c^i$ as:
\begin{equation}
\Delta P_k^i=\left\{
\begin{aligned}
c^i ( F_{s}^i, F_{l}^i, \Delta P_k^{i+1} ) & , & i=1,2,3,4 \\
c^i ( F_{s}^i, F_{l}^i ) & , & i=5
\end{aligned}
\right.
\label{offset}
\end{equation}
where $\Delta P_k^i = \{ \Delta p_k^i \}$. The deepest 5-th level $\Delta P_k^5$ is first computed. Since the misalignment between two features $F_{s}^i$ and $F_{l}^i$ of the deepest level is small \cite{gao2019dynamic}, the learning of offsets is relatively accurate and less challenging. We then pass the learned offsets to the upper levels to learn more precise offsets. This process is done level-by-level as a hierarchical refinement \cite{wang2019edvr, luo2021ebsr, chang2021low}.

After long-exposure features $F_l^1$-$F_l^5$ are aligned, we perform the feature fusion. In $i$-th level's Feature fusion block, the aligned features $F_{a}^i$ and short-exposure features $F_{s}^i$ are concatenated and then processed by a Residual block $r^i$ as:
\begin{equation}
F_{f}^i = r^i ( F_{s}^i, F_{a}^i ),
\label{fea_fusion}
\end{equation}
where the output features $F_{f}^5$ are connected to the first decoder layer, while $F_{f}^1$-$F_{f}^4$ serve as short-cut connections like in UNet \cite{ronneberger2015u}. Finally, we use a Residual block (including 4 residual layers) at the tail to further refine the features at the original resolution. The final output is added to $t^{\uparrow}$.

\noindent \textbf{Loss.} We first train DeblurNet and then train EnhanceNet. The L1 loss \cite{zhang2017beyond} is used for training them, as shown in Figure \ref{net1}. They are expressed as:
\begin{equation}
L_1^{DeblurNet} = \mathbb{E} [|| t - z^{\downarrow} ||_1 ], \ L_1^{EnhanceNet} = \mathbb{E} [|| y - z ||_1 ],
\label{loss}
\end{equation}
where $z^{\downarrow}$ is the average pooling downsampled result from the ground truth $z$ to match the resolution of $t$.

\begin{figure*}[t]
\centering
\includegraphics[width=\linewidth]{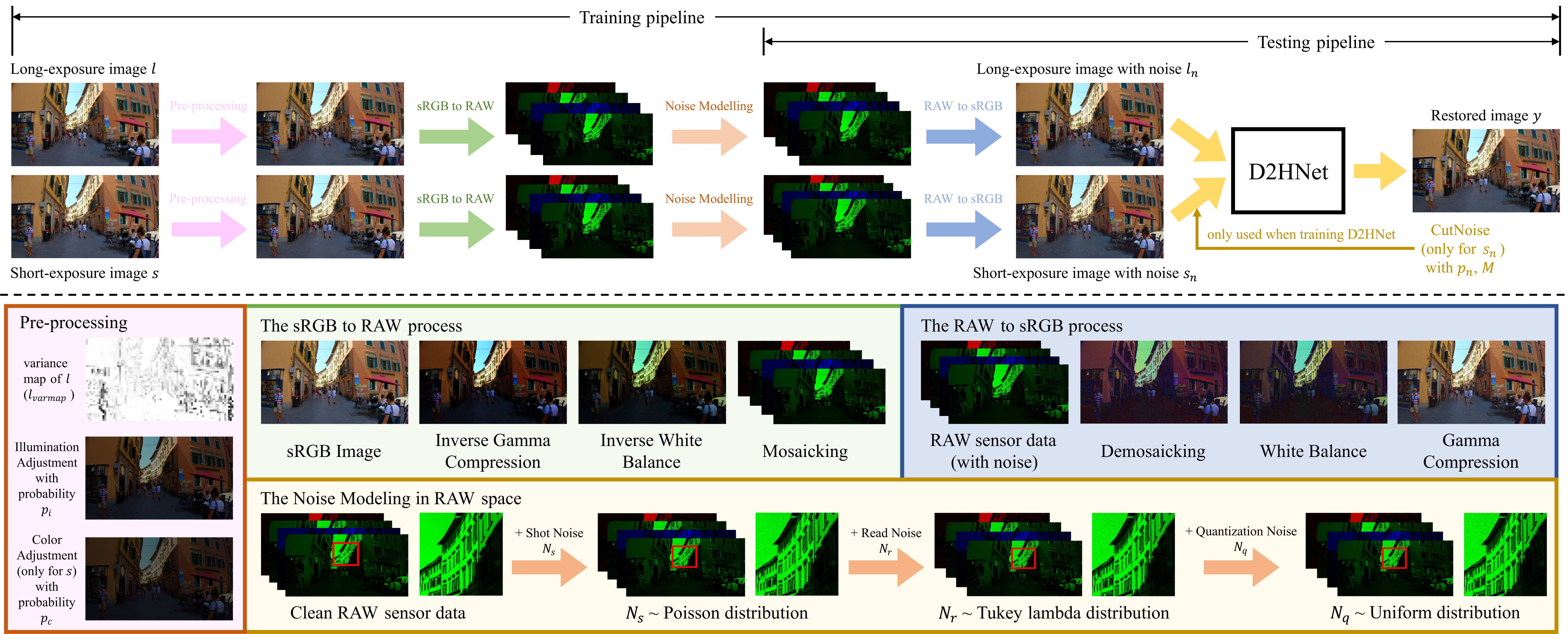}

\caption{Illustration of detailed procedures of training and testing pipelines (upper) and examples of every step of data processing procedures (lower).}
\label{pipeline}

\end{figure*}

\begin{figure}[t]
\centering
\includegraphics[width=120mm, height=33mm]{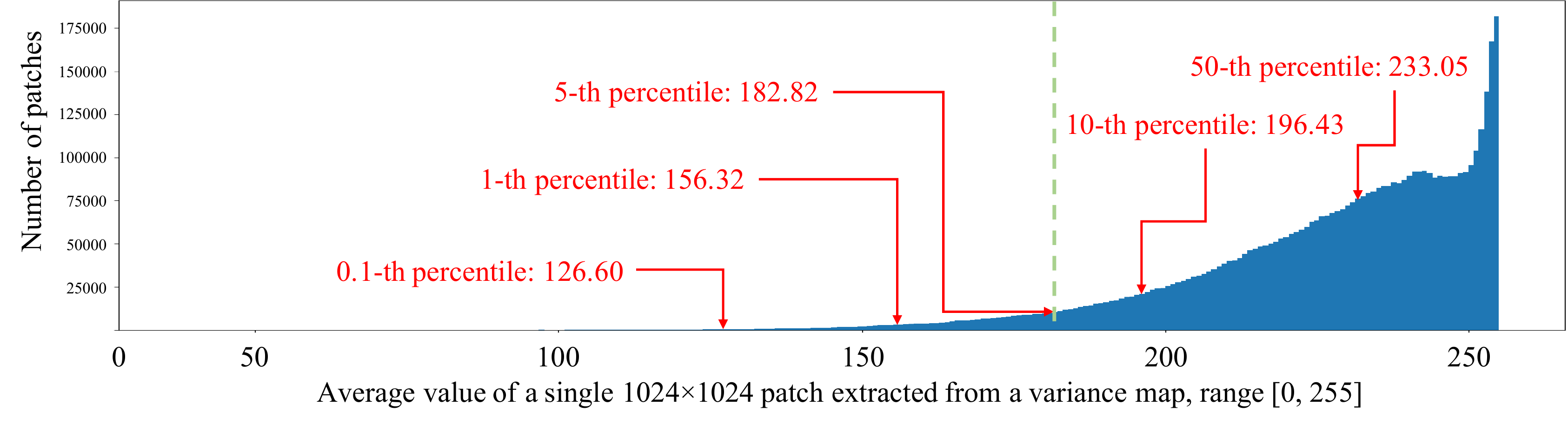}

\caption{Variance map patch distribution on D2-Dataset. The 5-th percentile point 182.82 is set as the threshold. If the average of a variance map patch is lower than 182.82, its corresponding long-exposure patch is regarded to be highly blurry.}
\label{distribution}

\end{figure}

\subsection{Data Processing}

The data processing is a key process for training the D2HNet, which includes VarmapSelection, Appearance Adjustment, Noise Modeling, and CutNoise, as shown in Figure \ref{pipeline}. It is designed to better model the real image distortion and balance the training data distribution.

\noindent \textbf{VarmapSelection.} It is a variance-map-based selection scheme to address the imbalanced blurriness issue and improve the robustness of D2HNet. Since most regions of the simulated long-exposure image $l$ are of low levels of blur, simply randomly choosing patches at the training causes loss fluctuation and ineffective deblurring ability for large motion. VarmapSelection scheme uses the variance map to represent the blur level and choose larger motion patches for training. The variance map is calculated from both $l$ and $l_{last}$ as:
\begin{equation}
l_{varmap} = {\rm min} ( {\rm Var} ( l ) ) / {\rm Var} ( l_{last} ) , 1 ),
\label{varmap_da}
\end{equation}
where ${\rm Var}$ computes the variance using a $k \times k$ window, with stride $k$. The resulted map is $1/k \times 1/k$ of the original size, so we upsample it with the nearest interpolation. According to the definition, a smaller value means higher levels of blur. One example variance map is shown in the pink rectangle of Figure \ref{pipeline}, where the blurrier regions in $l$ have clearer (darker) responses in $l_{varmap}$.

To determine whether a randomly chosen patch is of a large blur, we choose to define a threshold based on variance statistics on the training set. For each variance map, we randomly sample 1000 different squares of size 1024$\times$1024 and calculate the average variance value for each square. Then, we sort all values across the dataset and use the 5-th percentile point as the threshold, as shown in Figure \ref{distribution}. Afterward, we do sampling again, keep only the squares that have lower average variance values than the threshold and draw patches from the corresponding long-short-GT tuples. This process results in additional 9453 tuples of a strong blur. They are added to the original training set.

\noindent \textbf{Appearance Adjustment.} To simulate low-light image tuples, we apply Illumination Adjustment (IA) to lower the overall brightness. We also apply Color Adjustment (CA) to model the difference between long- and short-exposure images. An example is shown in the pink rectangle of Figure \ref{pipeline}. IA is done by an inverse gamma compression for long-short-GT tuples, as follows:
\begin{equation}
IA( u ) = {\rm max} ( u, \varepsilon )^{ g }, \ {\rm for}\ u \in \{s, l, z\},
\label{pre_processing}
\end{equation}
where $\varepsilon=10^{-8}$. The gamma value $g$ is randomly chosen from [1/0.6, 1/0.7, 1/0.75, 1/0.8, 1/0.9]. CA is achieved by a linear transform to disturb the overall color and brightness for only the short-exposure image $s$. It is defined as:
\begin{equation}
CA ( s ) = a \cdot s + b,
\label{pre_processing2}
\end{equation}
where $a$ and $b$ are sampled uniformly from [0.3, 0.6] and [0.001, 0.01], respectively.

\noindent \textbf{Noise Modeling.} We calibrate real smartphone noises in the RAW image space following \cite{wei2020physics} and then apply the noise simulation. Since our D2-dataset contains only sRGB images, we adopt a simple reverse ISP process \cite{brooks2019unprocessing} to convert them from sRGB to RAW. It includes an inverse gamma compression (as in Equation \ref{pre_processing} with $g = 2.2$), an inverse white balance which simply scales R and B channels by scalar $1/w_r, 1/w_b$ separately, and the mosaic to form Bayer pattern. Note white balance gain for the G channel ($w_g$) is fixed to 1, while $w_r$ and $w_b$ are sampled uniformly from [1.9, 2.4] and [1.5, 1.9], respectively. After noise modeling, we then convert it to sRGB by a forward ISP process with corresponding parameters for inversion. The noise model is calibrated on the \emph{Xiaomi Mi Note 10}, which we use to capture real photos.

\noindent \textbf{CutNoise.} To encourage the fusion and utilization of the short-exposure image, inspired by \cite{yoo2020rethinking}, we design the CutNoise scheme. It is performed after noise simulation on the short-exposure image which has stronger noise. CutNoise randomly selects a region and copies ground truth $z$ (i.e., $s_{first}$) to the corresponding position of $s_n$. The region itself can be any shape but we fix it to square for easy implementation. With CutNoise, D2HNet will not degenerate to use only the blurrier long-exposure input but is forced to learn to fuse information from the sharper short-exposure one, therefore generating sharper output.


\section{Experiment}

\subsection{Implementation Details}

Our training samples include original 5661 tuples of full-resolution images from the D2-Dataset and 9453 tuples of strong blurry patches selected by the VarmapSelection scheme. For DeblurNet, the input resolution is fixed to 512$\times$512 by average pooling. The epochs are 100 and the learning rate is initialized as $1 \times 10^{-4}$. For EnhanceNet, the input resolution is 256$\times$256 randomly cropped patches due to memory limit. The epochs are 150 and the learning rate is initialized as $5 \times 10^{-5}$. For both subnets, the learning rates are halved every 50 epochs. The batch size equals 2 and an epoch includes 5661 iterations, corresponding to the number of training tuples. The Adam optimizer \cite{kingma2014adam} with $\beta_1 = 0.5$ and $\beta_2 = 0.999$ is used. The probabilities of performing Illumination Adjustment, Color Adjustment, and CutNoise are set to 0.3, 0.5, and 0.3, respectively. The size of the CutNoise square is 120. We implement the D2HNet with PyTorch 1.1.0 and train it on 2 Titan Xp GPUs. It takes approximately 2 weeks to complete the optimization.

\begin{table}[t]
\begin{minipage}{0.58\linewidth}
\centering
\caption{Comparisons of D2HNet and other methods on D2-Dataset validation set by PSNR and SSIM \cite{wang2004image}. The \textcolor{red}{red} and \textcolor{blue}{blue} colors denote the best and second-best results, respectively.}
\label{sota_val}
\resizebox{59mm}{19mm}{
\begin{tabular}{lcccc}
\hline
\multirow{2}{*}{Method} & \multicolumn{2}{c}{1440p val data} & \multicolumn{2}{c}{2880p val data} \cr & PSNR & SSIM & PSNR & SSIM \\
\hline
DenseFuse \cite{li2018densefuse} & 32.90 & 0.9484 & 34.70 & 0.9637 \\
LSD2 \cite{mustaniemi2020lsd} & 33.20 & 0.9517 & 35.36 & 0.9675 \\
LSFNet \cite{chang2021low} & 33.87 & 0.9557 & 36.17 & 0.9715 \\
DeblurGAN \cite{kupyn2018deblurgan} & 33.80 & 0.9558 & 36.26 & 0.9701 \\
SGN \cite{gu2019self} & 33.87 & 0.9567 & 36.25 & 0.9720 \\
\hline
TP1 & 34.35 & \textcolor{blue}{0.9628} & 36.66 & \textcolor{blue}{0.9755} \\
TP2 & 34.26 & 0.9599 & 36.54 & 0.9733 \\
TP3 & \textcolor{blue}{34.41} & 0.9611 & \textcolor{blue}{36.70} & 0.9747 \\
\hline
D2HNet & \textcolor{red}{34.67} & \textcolor{red}{0.9639} & \textcolor{red}{36.85} & \textcolor{red}{0.9767} \\
\hline
\end{tabular}
}
\end{minipage}
\hspace{1mm}
\begin{minipage}{0.4\linewidth}  
\centering
\caption{The results of the human perceptual study on real photos for the D2HNet and other pipelines, given by preference rates (PR) for D2HNet over all the votes.}
\label{sota_pr}
\resizebox{44mm}{14.5mm}{
\begin{tabular}{cc}
\hline
Method & PR \\
\hline
D2HNet $>$ DenseFuse, & \multirow{3}{*}{86.07\%} \\
LSD2, LSFNet & \\
DeblurGAN, SGN & \\
\hline
D2HNet $>$ TP1 & 82.50\% \\
D2HNet $>$ TP2 & 81.07\% \\
D2HNet $>$ TP3 & 79.28\% \\
\hline
\end{tabular}
}
\end{minipage}

\end{table}

\begin{figure*}[t]
\centering
\includegraphics[width=\linewidth]{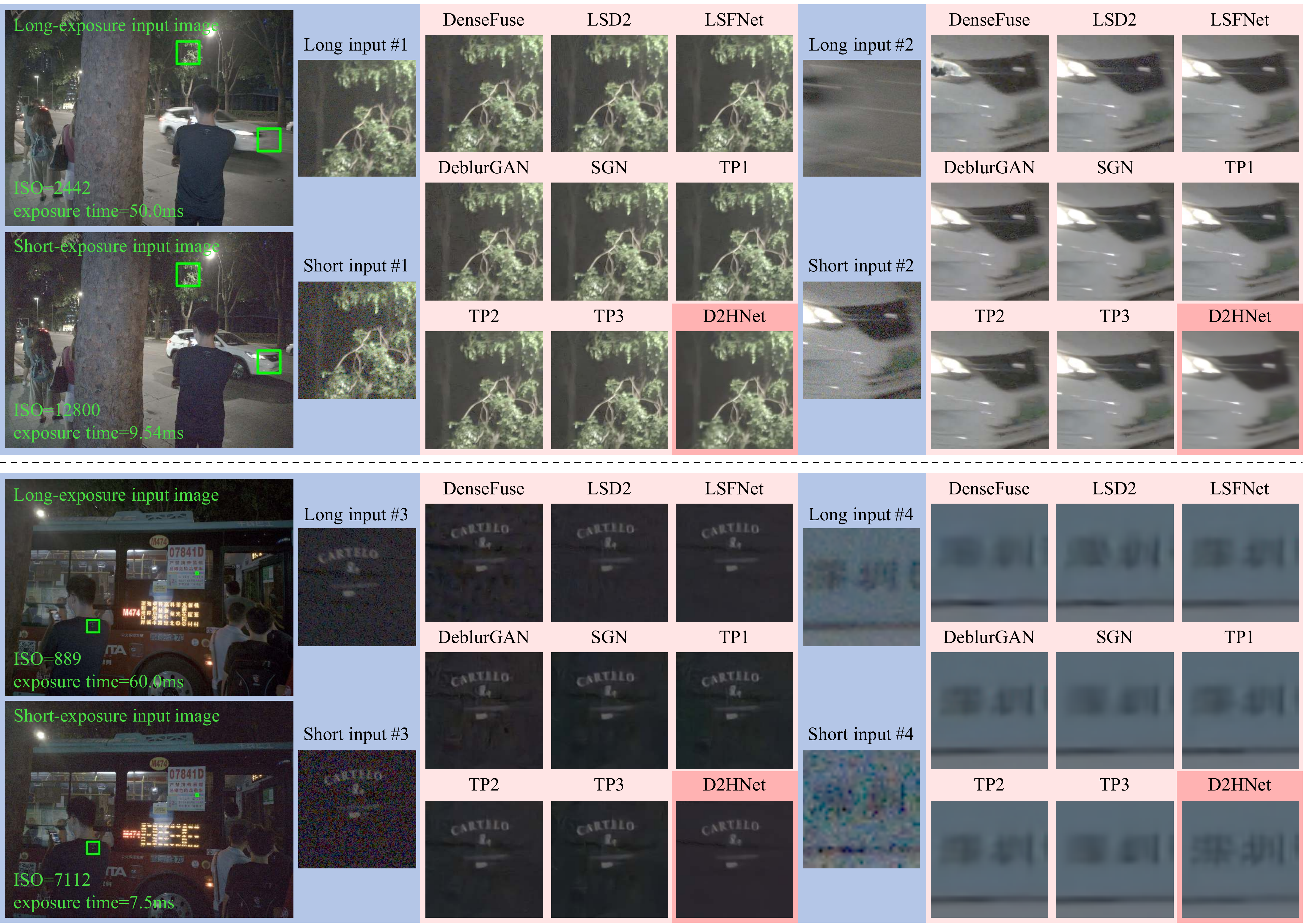}

\caption{Visual comparisons of the proposed D2HNet with other methods. More results on both real photos and validation images are in the supplementary material.}
\label{sota}

\end{figure*}

\subsection{Long-short Fusion Method Experiments}

We compare the image restoration quality of D2HNet and other recent works with similar target, DenseFuse \cite{li2018densefuse}, LSD2 \cite{mustaniemi2020lsd}, and  LSFNet \cite{chang2021low}, or with SOTA performance in either denoising or deblurring, SGN \cite{gu2019self}, DeblurGAN \cite{kupyn2018deblurgan} (see more in Section 5.3). To fit the dual inputs, SGN's and DeblurGAN's input layers are changed to receive two images. In addition, we define three more two-phase pipelines for a more comprehensive evaluation: 1) image denoising by SGN + long-short fusion by SGN (denoted as TP1); 2) image deblurring by DeblurGAN + long-short fusion by SGN (denoted as TP2); 3) long-short fusion by SGN + long-short fusion refinement by SGN (the same workflow as D2HNet, denoted as TP3). The same data processing schemes are applied to other methods.

We illustrate the generated samples on real photos in Figure \ref{sota}. From image pairs $\sharp$1 and $\sharp$3, the black backgrounds of D2HNet results are cleaner than other methods, e.g., obvious artifacts in results of DenseFuse, DeblurGAN, and TP1-TP3. It demonstrates that D2HNet has a better denoising ability for inputs. For image pair $\sharp$2, D2HNet can generate a clean and sharp result from extreme blurry inputs, while maintaining the denoising ability of dark regions; however, there lie in artifacts in the dark regions of others. From image pairs $\sharp$3 and $\sharp$4, we can see D2HNet has better edge preservation ability compared with others, e.g., letters and Chinese characters are sharper and cleaner.

\begin{table}[t]
\begin{minipage}{0.49\linewidth}
\centering
\caption{Comparisons of D2HNet and other single image denoising methods.}
\label{denoise_val}
\resizebox{59mm}{21mm}{
\begin{tabular}{lcccc}
\hline
\multirow{2}{*}{Method} & \multicolumn{2}{c}{1440p val data} & \multicolumn{2}{c}{2880p val data} \cr & PSNR & SSIM & PSNR & SSIM \\
\hline
DnCNN \cite{zhang2017beyond} & 32.20 & 0.9192 & 33.61 & 0.9265 \\
MemNet \cite{tai2017memnet} & 33.74 & 0.9517 & 35.73 & 0.9644 \\
MWCNN \cite{liu2018multi} & 32.47 & 0.9372 & 34.71 & 0.9554 \\
SGN \cite{gu2019self} & 33.94 & \textcolor{blue}{0.9576} & \textcolor{blue}{36.42} & \textcolor{blue}{0.9713} \\
RIDNet \cite{anwar2019real} & 33.29 & 0.9462 & 35.55 & 0.9621 \\
MIRNet \cite{zamir2020learning} & 33.98 & 0.9565 & 36.36 & 0.9708 \\
REDI \cite{lamba2021restoring} & 28.60 & 0.8964 & 31.54 & 0.9431 \\
DeamNet \cite{ren2021adaptive} & 33.78 & 0.9531 & 36.26 & 0.9685 \\
MPRNet \cite{zamir2021multi} & \textcolor{blue}{34.00} & 0.9568 & 36.25 & 0.9712 \\
\hline
D2HNet & \textcolor{red}{34.67} & \textcolor{red}{0.9639} & \textcolor{red}{36.85} & \textcolor{red}{0.9767} \\
\hline
\end{tabular}
}
\end{minipage}
\hspace{1mm}
\begin{minipage}{0.49\linewidth}
\centering
\caption{Comparisons of D2HNet and other single image deblurring methods.}
\label{deblur_val}
\resizebox{59mm}{21mm}{
\begin{tabular}{lcccc}
\hline
\multirow{2}{*}{Method} & \multicolumn{2}{c}{1440p val data} & \multicolumn{2}{c}{2880p val data} \cr & PSNR & SSIM & PSNR & SSIM \\
\hline
\scriptsize DeepDeblur \cite{nah2017deep} & 23.51 & 0.8252 & 23.80 & 0.8731 \\
SRN \cite{tao2018scale} & 23.99 & 0.8363 & 24.11 & \textcolor{blue}{0.8780} \\
\scriptsize DeblurGAN \cite{kupyn2018deblurgan} & \textcolor{blue}{24.23} & \textcolor{blue}{0.8399} & \textcolor{blue}{24.13} & 0.8749 \\
\scriptsize DeblurGANv2 \cite{kupyn2019deblurgan} & 23.88 & 0.8059 & 23.67 & 0.8359 \\
DMPHN \cite{zhang2019deep} & 21.73 & 0.7807 & 22.38 & 0.8447 \\
MPRNet \cite{zamir2021multi} & 22.97 & 0.8072 & 22.61 & 0.8438 \\
HINet \cite{chen2021hinet} & 22.39 & 0.7586 & 21.93 & 0.7879 \\
\scriptsize MIMOUNet \cite{cho2021rethinking} & 21.11 & 0.7756 & 21.19 & 0.8355 \\
\scriptsize MIMOUNet++ \cite{cho2021rethinking} & 21.10 & 0.7753 & 21.25 & 0.8373 \\
\hline
D2HNet & \textcolor{red}{34.67} & \textcolor{red}{0.9639} & \textcolor{red}{36.85} & \textcolor{red}{0.9767} \\
\hline
\end{tabular}
}
\end{minipage}

\end{table}

\begin{figure*}[t]
\centering
\includegraphics[width=\linewidth]{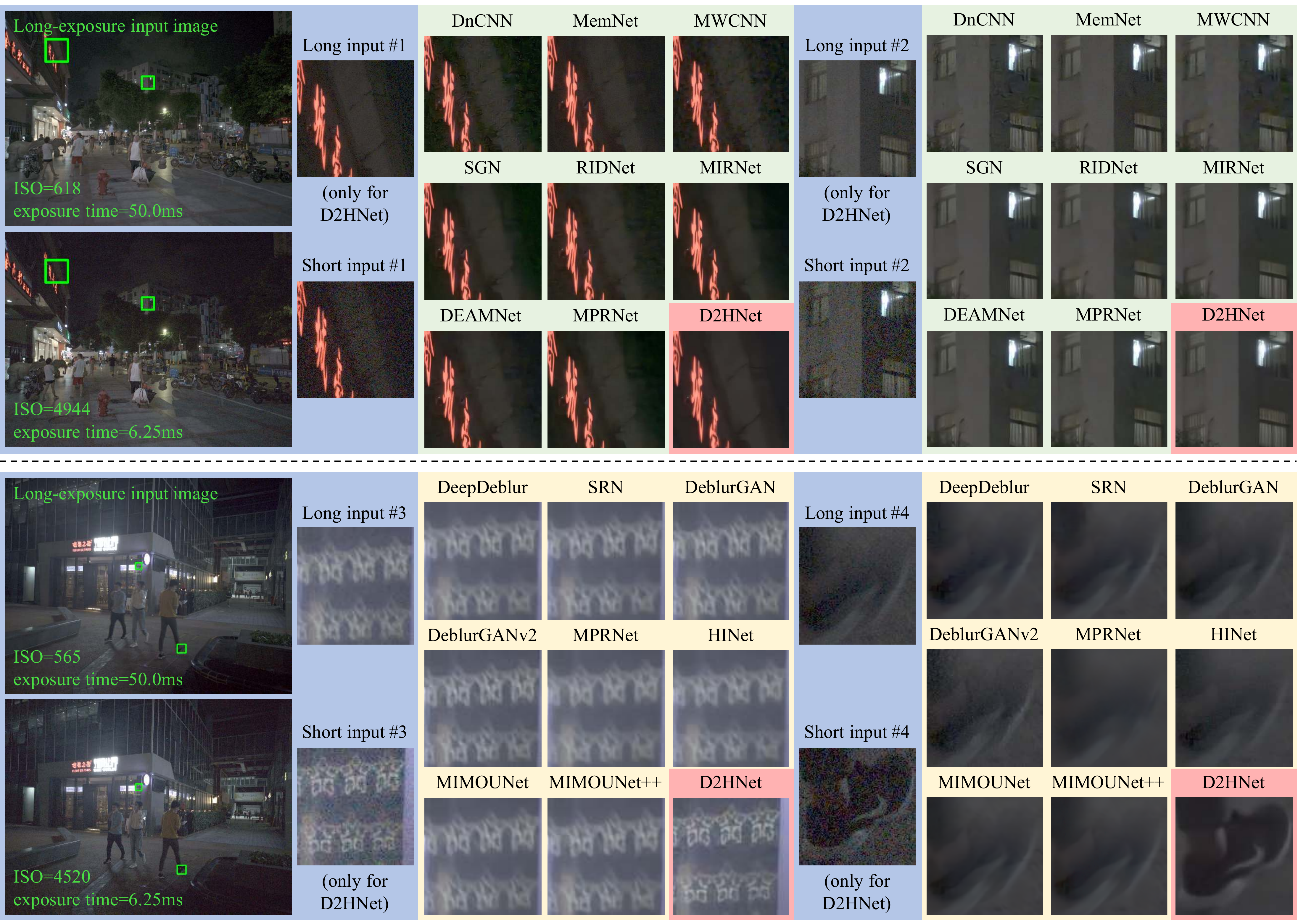}

\caption{Visual comparisons of the proposed D2HNet with single image denoising methods (upper $\sharp$1 and $\sharp$2) and single image deblurring methods (lower $\sharp$3 and $\sharp$4).}
\label{sota_de2}

\end{figure*}

The quantitative analysis is concluded in Table \ref{sota_val}. Compared with other single-phase methods, D2HNet obtains 0.80$\sim$1.77db PSNR gain on 1440p. It also outperforms the simple concatenated methods (TP1-TP3) on both 1440p and 2880p, which demonstrates that D2HNet is more robust to different input resolutions. Since there is no ground truth for real photos, we conduct a human perceptual study on the results generated from different methods and there are 10 observers. In each comparison, a user is presented with a pair of restored images side by side of a shuffled sequence. Then, the user chooses one result that produces cleaner and sharper images than others. The preference rates (PRs) are concluded in Table \ref{sota_pr}, where there are 79.28\%$\sim$86.07\% votes for D2HNet. The majority of users thought that D2HNet achieves higher image quality than compared methods. It demonstrates that D2HNet recovers images with better details and textures and well addresses the domain gap issue.

\subsection{Single-image Denoising and Deblurring Method Experiments}

We compare D2HNet and SOTA image denoising \cite{zhang2017beyond, tai2017memnet, liu2018multi, gu2019self, anwar2019real, zamir2020learning, zamir2021multi, lamba2021restoring, ren2021adaptive} and deblurring \cite{nah2017deep, tao2018scale, kupyn2018deblurgan, kupyn2019deblurgan, zhang2019deep, zamir2021multi, cho2021rethinking, chen2021hinet} methods. Short-exposure images serve as inputs for denoising methods and $s_{first}$ is ground truth. Long-exposure images serve as inputs for deblurring methods and $l_{last}$ is ground truth.

We illustrate the generated samples on real photos in Figure \ref{sota_de2}. From $\sharp$1 and $\sharp$2, single image denoising methods cannot restore details of the roof ($\sharp$1) and the textures of curtains ($\sharp$2). However, D2HNet produces richer details since it fuses the information from the long-exposure input, where the textures are more distinguishable than the highly noisy short-exposure input. From $\sharp$3 and $\sharp$4, single image deblurring methods cannot recover either small blur or severe blur. The superiority of D2HNet comes from two reasons. On one hand, although the other methods estimate motion fields from a single long-exposure input, D2HNet utilizes the position information from the short-exposure input to guide the deblurring. On the other hand, a domain gap exists between training and testing data. Without proper handling, these methods degrade to mainly removing noises when encountering very large blurs in the testing images. Whereas, our architecture involves the DeblurNet which operates on a fixed resolution to better generalize on large blur. We also report the quantitative performance of all methods on the validation set in Table \ref{denoise_val} and \ref{deblur_val}. Compared with single-image-based methods, D2HNet obtains giant increases on both metrics since it fuses more information from both long- and short-exposure inputs.

\subsection{Ablation Study}

\begin{figure*}[t]
\centering
\includegraphics[width=120mm, height=45mm]{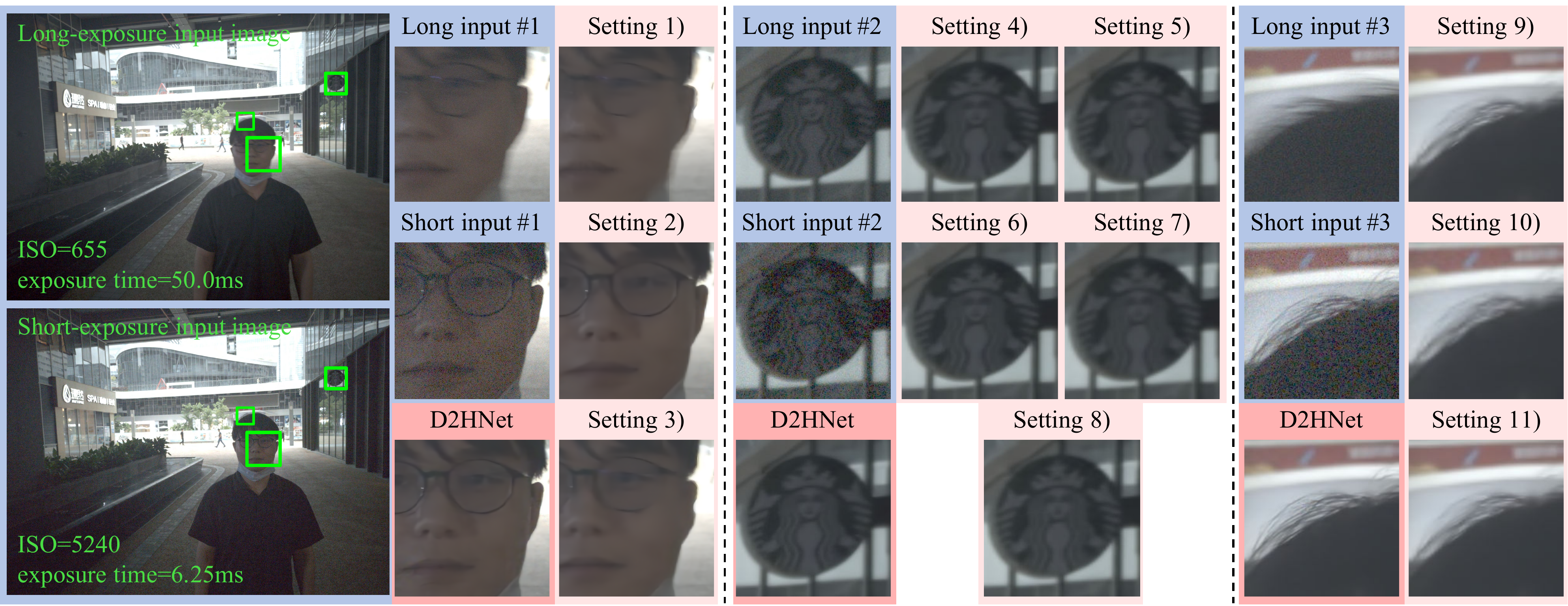}

\caption{Visual comparisons of D2HNet ablation study.}
\label{ablation}

\end{figure*}

We conduct the ablation study for the D2HNet, where the benchmark results are concluded in Table \ref{ab_val} and visual results are illustrated in Figure \ref{ablation}\footnote[1]{We thank Chao Wang in the SenseTime Research for helping capture the image.}. The analysis for different ablation study items is as follows:

\noindent \textbf{Training Strategy.} Dual inputs are significant for D2HNet to get more performance gain. Compared with only using long- or short-exposure input (settings 1) and 2)), two inputs improve PSNR by 9.68dB and 0.59dB, respectively. Aligning the long-exposure input with short-exposure input (i.e., $s_{first}$ as GT) also helps transfer textures from long-exposure input, which brings 4.51dB gain compared with $l_{last}$ as GT (setting 3)). We can also see settings 1-3) cannot recover the details and remove artifacts (e.g., the face contour and eyes in $\sharp$1).

\noindent \textbf{Network Components.} Alignment block makes the D2HNet better fuses features from the long-exposure input. In setting 4), we replace deformable convolutions with ordinary convolutions, forcing the network to apply rigid filters at all the spatial locations in the features, which brings a decrease of 0.24dB. In setting 5), we remove all Alignment and Feature fusion blocks, leading to a notable performance decrease of EnhanceNet (1.4dB) since the hierarchical information is excluded. From settings 6) and 7), the tail Residual block brings 0.5dB gain, while the full EnhanceNet brings 4.19dB gain since it learns rich textures and details. In addition, settings 4-7) produce blurry outputs and vague details (i.e., the billboard in $\sharp$2), which show the importance of every component.

\noindent \textbf{Data Processing Schemes.} VarmapSelection balances the training data distribution, where D2HNet better generalizes to blurry or misaligned long-exposure inputs and learns to extract textures from them, e.g., D2HNet produces sharper results than setting 8) in $\sharp 2$. Illumination Adjustment generates more low-brightness training images, helping the D2HNet obtain better performance in dark regions. Color Adjustment and CutNoise balance the usage of long-short inputs, encouraging sharper results. As shown in $\sharp 3$, if dropping each of them (settings 9-11)), the network cannot recover clear details of the black hair. Also, settings 8-11) result in the decreases of PSNR by 0.43dB, 0.36dB, 0.09dB, and 0.25dB, respectively. It is obvious that every data processing scheme is significant.

\begin{table}[t]
\begin{center}
\caption{Comparisons of D2HNet and ablation settings on 1440p validation data.}
\label{ab_val}

\resizebox{\linewidth}{11.5mm}
{
\begin{tabular}{lcc|lcc}
\hline
Ablation Study Setting & PSNR & SSIM & Ablation Study Setting & PSNR & SSIM \\
\hline
1) Only long input, $l_{last}$ as ground truth & 24.99 & 0.8610 & 7) w/o EnhanceNet (only DeblurNet) & 30.48 & 0.9259 \\
2) Only short input & 34.08 & 0.9579 & 8) w/o VarmapSelection & 34.24 & 0.9604 \\
3) Long-short inputs, $l_{last}$ as ground truth & 30.16 & 0.9293 & 9) w/o Illumination Adjustment & 34.31 & 0.9596 \\
4) Replacing EnhanceNet Alignment block & 34.43 & 0.9610 & 10) w/o Color Adjustment & 34.58 & 0.9620 \\
5) w/o EnhanceNet feature-level short-cuts & 33.27 & 0.9530 & 11) w/o CutNoise & 34.42 & 0.9616 \\
6) w/o EnhanceNet tail Residual block & 34.17 & 0.9602 & D2HNet (full) & \textcolor{red}{34.67} & \textcolor{red}{0.9639} \\
\hline
\end{tabular}
}
\end{center}

\end{table}

\section{Conclusion}

In this paper, we present a D2HNet framework for robust night image restoration based on long- and short-exposure inputs. It deblurs and restores sharp outputs from the long-exposure image under the guidance of the short-exposure image to obtain accurate colors, trivial noises, and sharp edges. It includes two sequential subnets: DeblurNet to remove blur on a fixed size and EnhanceNet to refine and sharpen the output of DeblurNet. For training, we synthesize the D2-Dataset including 6853 high-quality image tuples with multiple types and levels of blur. We propose a VarmapSelection scheme to generate highly blurry patches and assist the convergence of D2HNet. We also use a CutNoise scheme to enhance textures and details by enforcing D2HNet to learn how and where to deblur. For evaluation, we compare the proposed D2HNet with SOTA long-short fusion methods, and single image denoising and deblurring methods on the D2-Dataset validation set and real-world photos. The experimental results on both validation set and real-world photos show better performance achieved by the D2HNet.

\clearpage
%
%
\bibliographystyle{splncs04}
\bibliography{egbib}

\appendix

\section*{Supplementary Material}

\section{More Results on Captured Real Images}

\begin{figure*}[t]
\centering
\includegraphics[width=\linewidth]{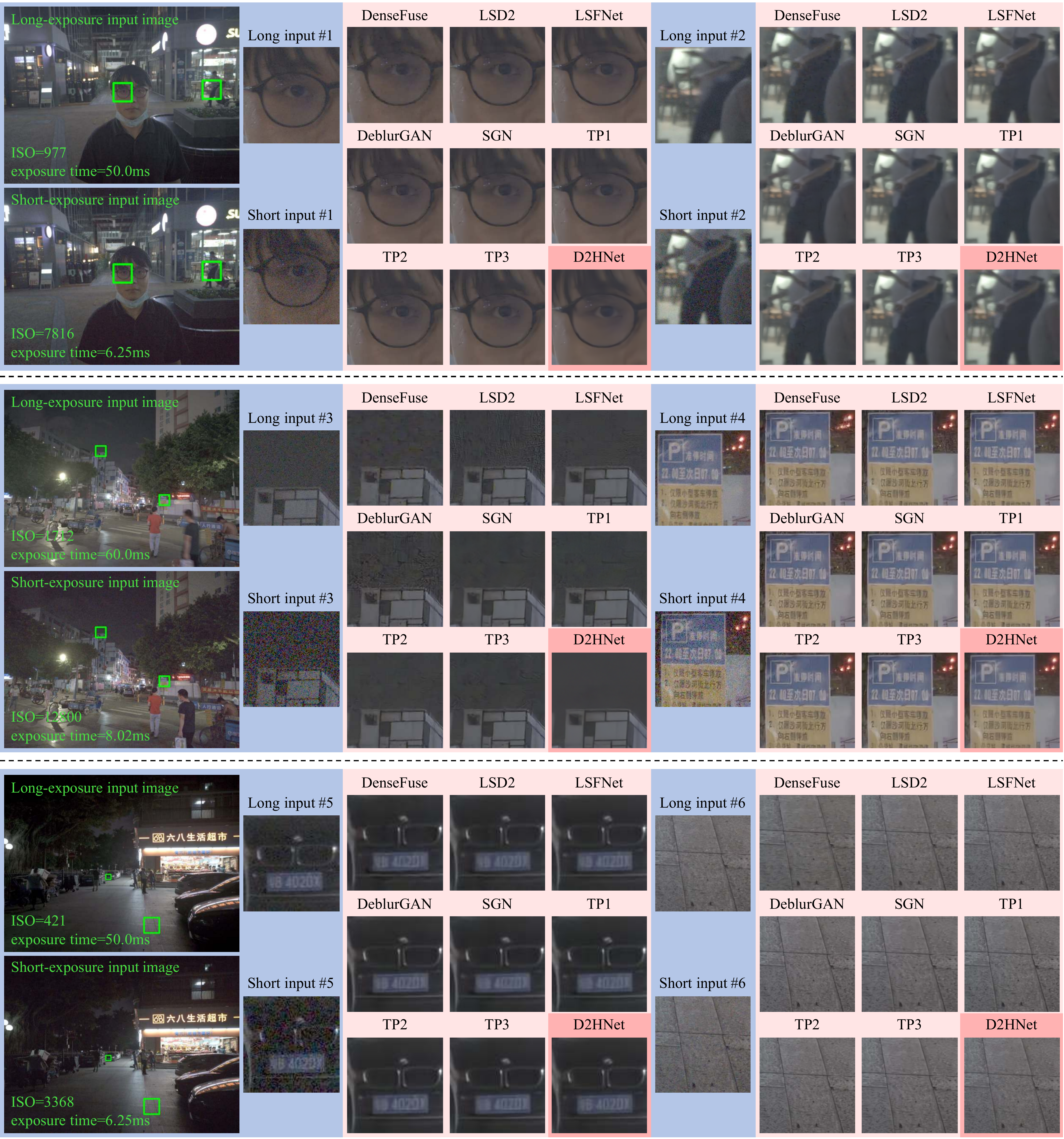}

\vspace{-0mm}

\caption{Visual comparisons of the proposed D2HNet with other methods on real photos. From $\sharp1$ and $\sharp2$, there are no visual artifacts of D2HNet, while there are obvious artifacts for other methods. From $\sharp3$, there are very obvious remaining noises in the dark sky of other methods, while the D2HNet output is much cleaner. From $\sharp4$ to $\sharp6$, we observe that D2HNet can well recover textures and remove artifacts simultaneously when there are a lot of details in the input images (especially in the long-exposure inputs). For instance, the Chinese characters in $\sharp4$ of D2HNet are cleaner and clearer than other methods; the letters and numbers ``B 4020X'' in $\sharp5$ of D2HNet are more distinguishable than other methods; D2HNet better learns the textures from $\sharp6$ inputs.}
\label{sota1}

\vspace{-0mm}

\end{figure*}

We show more visual results of D2HNet and SOTA methods on real images in Figure \ref{sota1}, which are captured with \emph{Xiaomi Mi Note 10} smartphone. The texture learning ability, denoising quality, and artifact removal performance of the proposed D2HNet are all better than SOTA methods. The more detailed analysis is in the captions.

\section{More Results on Validation Set}

\begin{figure*}[t]
\centering
\includegraphics[width=\linewidth]{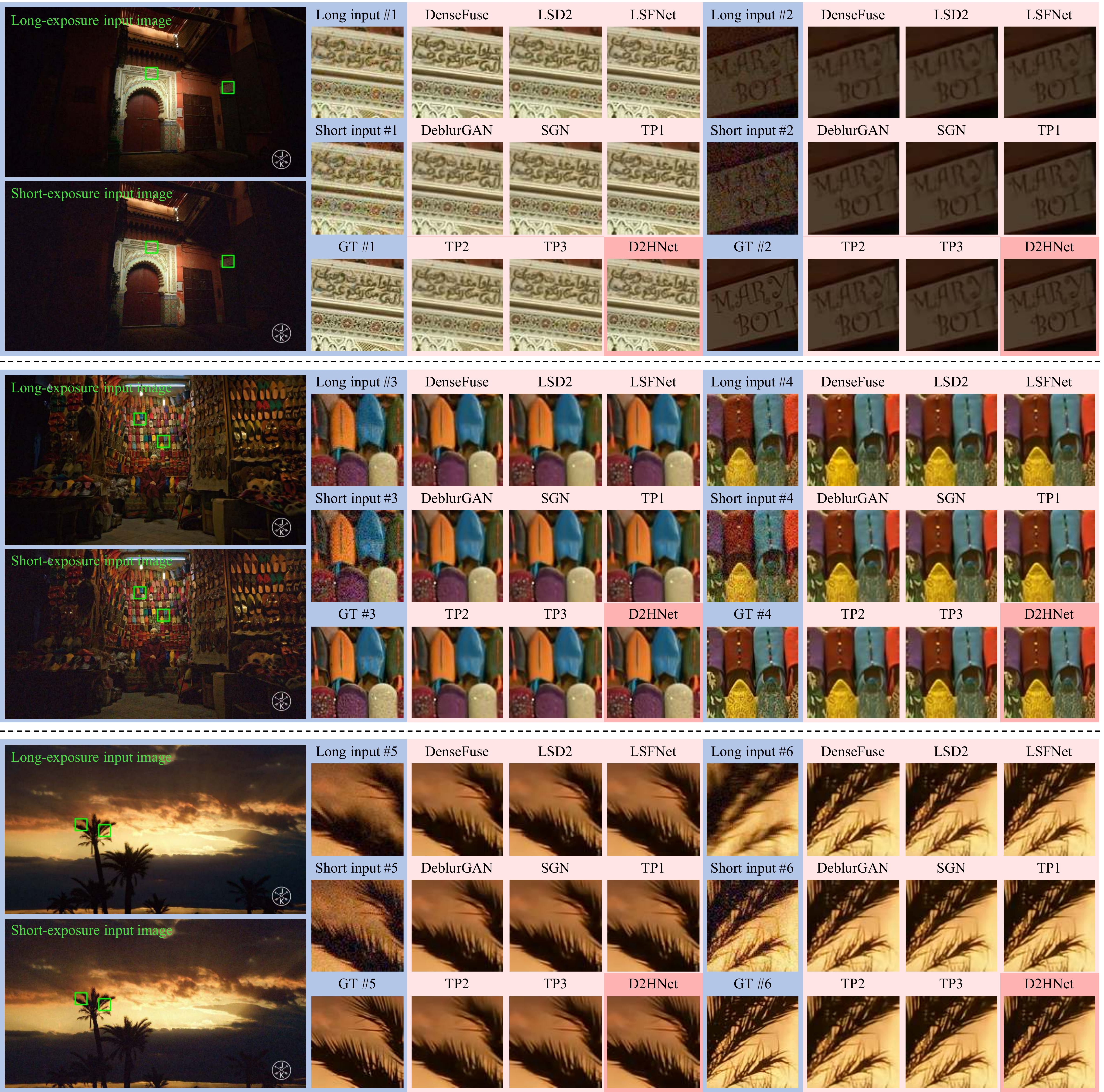}

\vspace{-0mm}

\caption{Visual comparisons of the proposed D2HNet with other methods on D2-Dataset 1440p validation set. Note that, we also show the ground truth since the experiments are performed on validation set. From $\sharp1$, $\sharp3$, and $\sharp4$, textures of the painting and shoes in D2HNet are clearer than other methods (please compare the details of different results based on ground truth, i.e., GT $\sharp1$, GT $\sharp3$, and GT $\sharp4$). From $\sharp2$, the letters of D2HNet are more distinguishable than other methods, e.g., the edges and clarity. From $\sharp5$ and $\sharp6$, the edges of D2HNet results are better than other methods.}
\label{val1}

\vspace{-0mm}

\end{figure*}

\begin{figure*}[t]
\centering
\includegraphics[width=\linewidth]{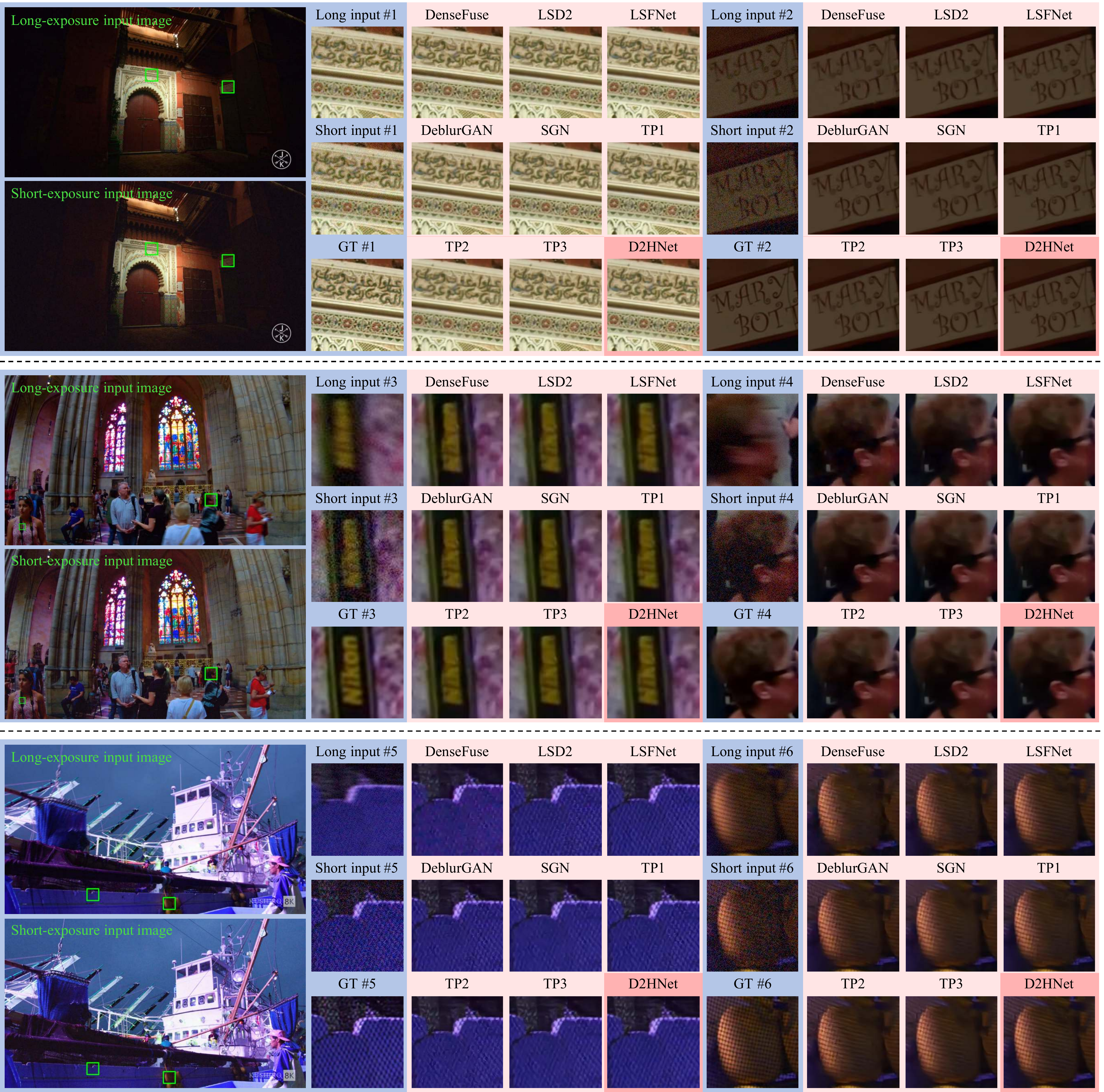}

\vspace{-0mm}

\caption{Visual comparisons of the proposed D2HNet with other methods on D2-Dataset 2880p validation set. Note that, we also show the ground truth since the experiments are performed on validation set. The input examples $\sharp1$ and $\sharp2$ are at the same relative positions to Figure \ref{val1}, but with different image resolutions. The textures and edges of D2HNet results are better than other methods. The proposed D2HNet performs well on both 1440p and 2880p validation images, which demonstrates that D2HNet has the ability to address the domain gap issue.}
\label{val2}

\vspace{-0mm}

\end{figure*}

We show more visual results of D2HNet and SOTA methods on the validation set of the collected D2-Dataset. The results on 1440p data and 2880p data are shown in Figure \ref{val1} and Figure \ref{val2}, respectively. The D2HNet produces more distinguishable details and achieves better deblurring quality. It also achieves consistent and better performance on different image resolutions.

\section{Burst-image Method Experiments}

\begin{figure*}[t]
\centering
\includegraphics[width=\linewidth]{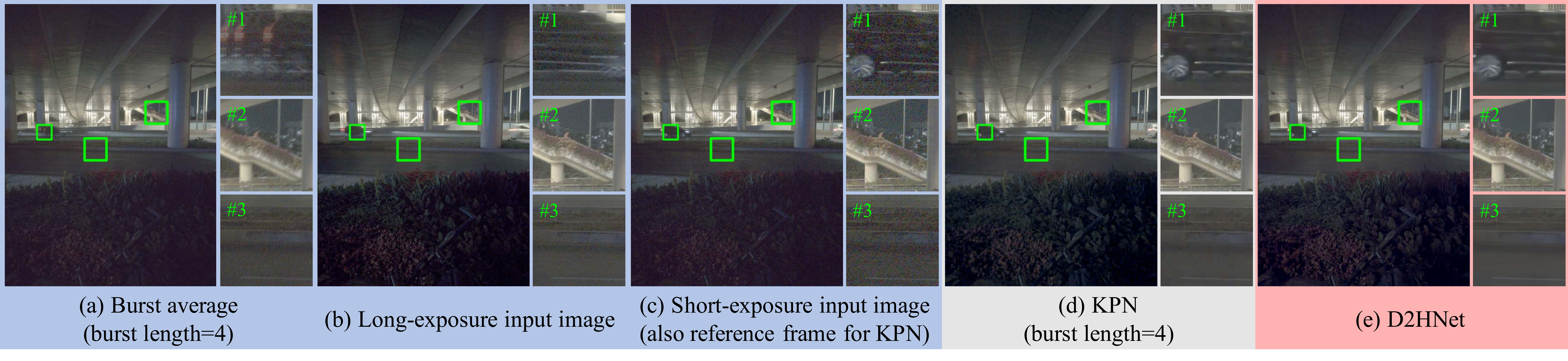}

\caption{Visual comparisons of the proposed D2HNet with KPN. The input long-short pairs and 4-frame images are captured by the same smartphone and in the same scene.}
\label{sota_kpn}

\end{figure*}

We compare D2HNet with a burst-image denoising method KPN \cite{mildenhall2018burst}. The training set of KPN is also generated from the same video source of D2-Dataset and 4 successive short-exposure images are synthesized by a similar process used in D2-Dataset, then augmented with the same noise parameters as D2HNet. The results are shown in Figure \ref{sota_kpn}, where D2HNet produces richer textures (e.g., flowers in $\sharp$2) and has fewer visual artifacts (e.g., black car in $\sharp$1 and dark road in $\sharp$3) than KPN. Since KPN defines a fixed size of output convolutional kernels, it is not flexible to image resolutions larger than training images, i.e., it cannot address the domain gap issue. In addition, burst capturing with 4 shots takes more time than 2 shots due to hardware constraints. And more shots introduce more misalignment issues. Hence our D2HNet framework is more favorable.

\section{More Results Related to Domain Gap}

\begin{figure*}[t]
\centering
\includegraphics[width=\linewidth]{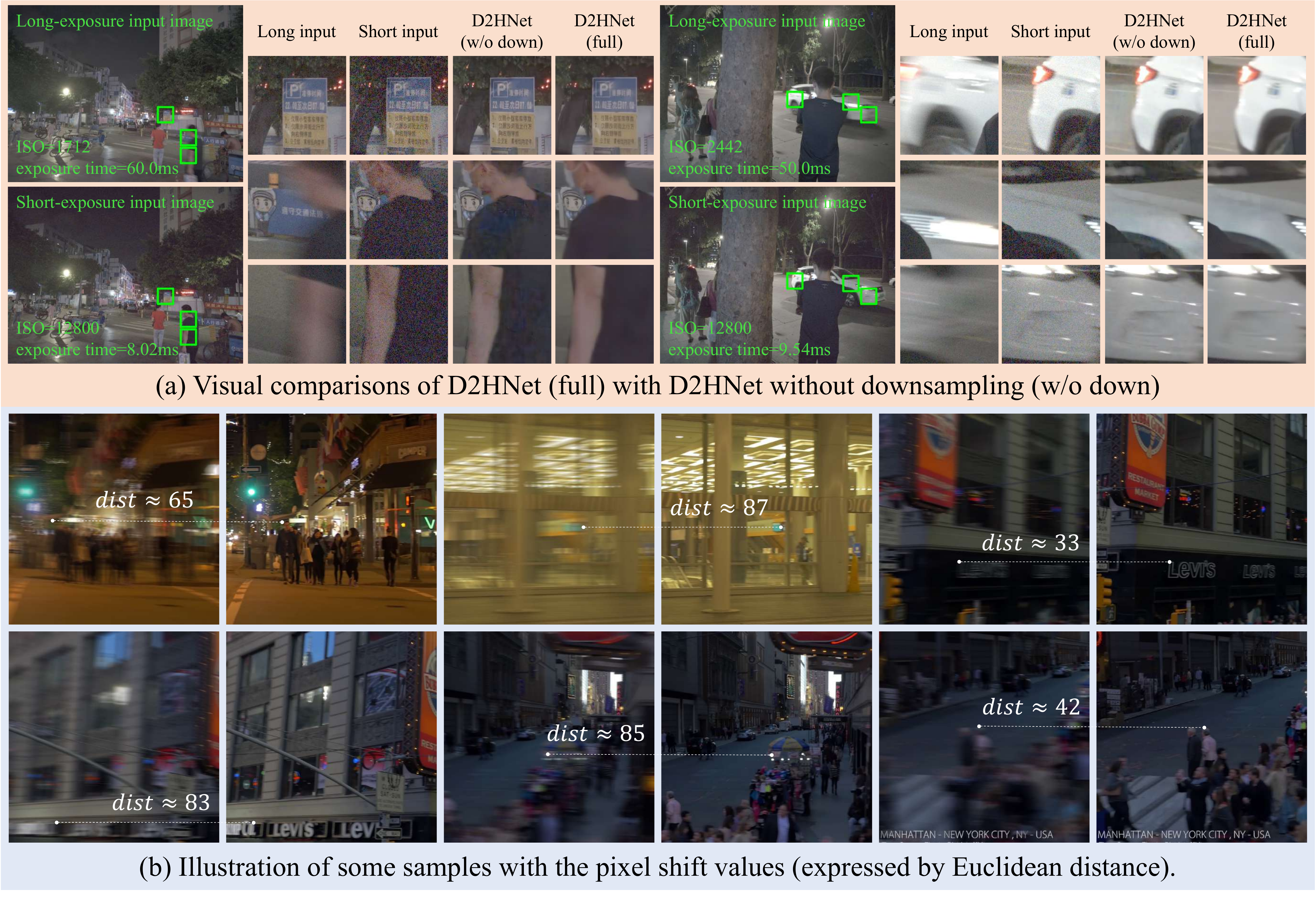}

\vspace{-0mm}

\caption{(a) Visual comparisons of the proposed D2HNet with the same architecture but without downsampling. We select some patches from the highly blurry areas in the long-exposure input. There are obvious artifacts of the D2HNet (w/o down) results, while much fewer artifacts are in D2HNet (full) results. Since D2HNet (w/o down) does not consider the domain gap issue, it cannot handle real-world inputs with larger pixel shifts than training images. Therefore, it simply copies the pixels of the long-exposure input to the output, e.g., there are many blue pixels on the black T-shirt of D2HNet (w/o down) results; (b) Illustration of pixel shift values of some training long- and short-exposure image pairs. The image pairs are selected from 9453 highly blurry tuples, which are obtained by the VarmapSelection scheme.}
\label{sota3}

\vspace{-0mm}

\end{figure*}

The domain gap in the task means differences between synthetic training images and real-world photos, e.g., blur area and resolution between them. To further demonstrate that D2HNet addresses the domain gap issue, we add an experiment setting that uses D2HNet architecture but does not perform downsampling for the input images of DeblurNet. The visual comparisons are shown in Figure \ref{sota3}. We observe that the pixel shifts of most highly blurry tuples are in the range of [40, 100], where some samples are shown in Figure \ref{sota3} (b). Since D2HNet architecture without downsampling only sees a maximum pixel shift of approximately 100, while the pixel shifts of the input pairs shown in Figure \ref{sota3} (a) are much larger than 100 (e.g., larger than 150 for the black T-shirt patch), it cannot handle such cases. Therefore, there are obvious artifacts in the results.

\section{Illustration of Data Acquisition}

We synthesize a D2-Dataset for training and benchmarking. There are three steps of the data synthesis pipeline, where the details are shown in Figure \ref{data} (a). For the data synthesis pipeline for training the burst-image denoising method, the details are shown in Figure \ref{data} (b). We also show some long- and short-exposure image pairs in Figure \ref{data} (c).

\section{Illustration of Data Processing Schemes}

\begin{figure*}[t]
\centering
\includegraphics[width=\linewidth]{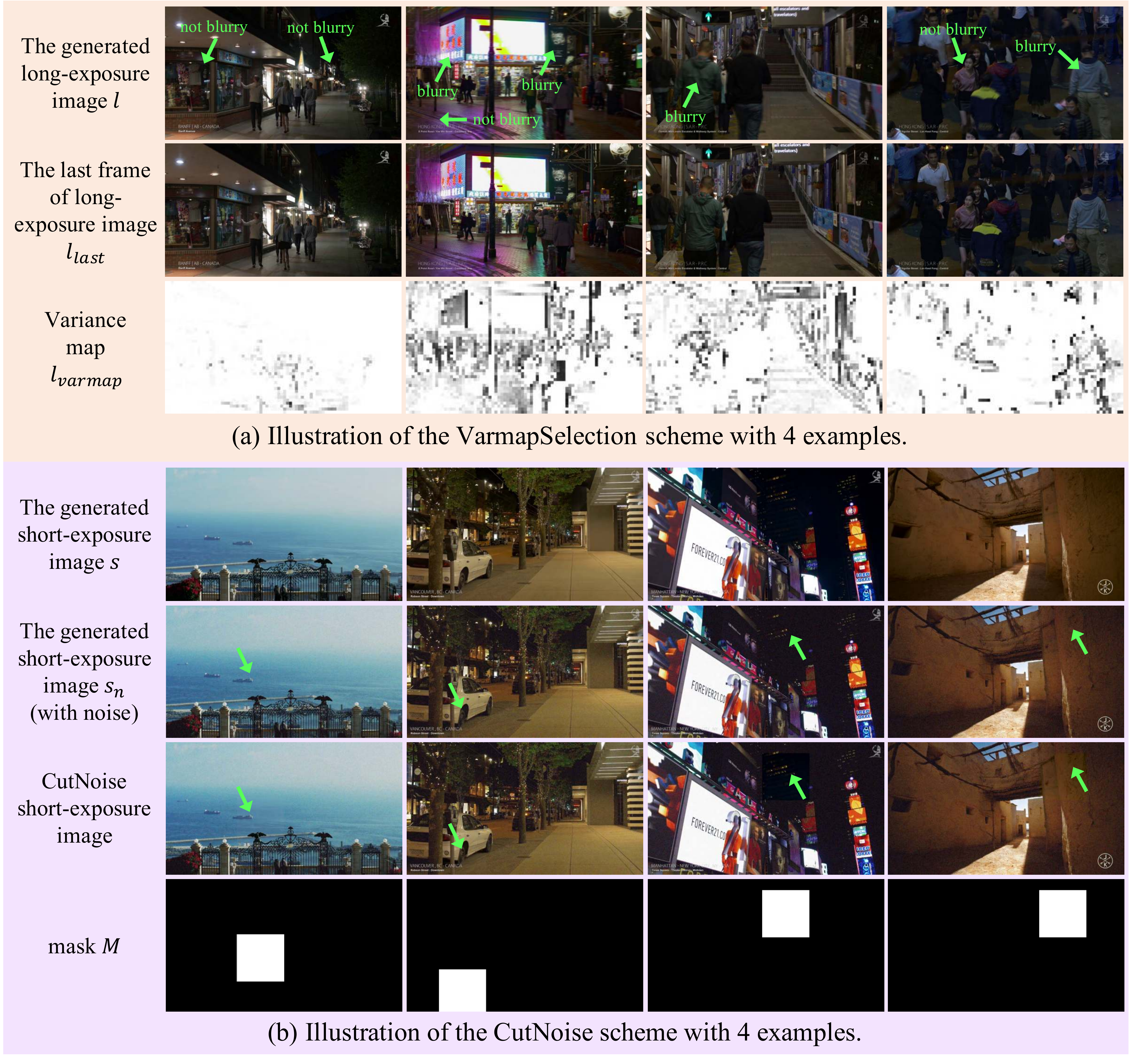}

\vspace{-0mm}

\caption{Illustration of the VarmapSelection and CutNoise schemes. The variance maps can reflect the blurry regions or regions with large motions, e.g., the dark regions in $l_{varmap}$. The VarmapSelection is effective and robust to select blurrier training patches from the whole dataset. The CutNoise can be expressed as $s_n^{CutNoise} = M \odot s_{first} + (\mathbbm{1} - M) \odot s_n $, where $\odot$ denotes matrix dot product and $\mathbbm{1}$ is an all-1 matrix with the same dimension of the binary mask $M$.}
\label{dp}

\vspace{-0mm}

\end{figure*}

To further visualize the effectiveness of VarmapSelection and CutNoise schemes, we show 4 examples in Figure \ref{dp}. The variance maps of VarmapSelection can well represent the regional blur degree; therefore, it helps select blurry patches at the training. It makes the D2HNet better generalize to blurry long-exposure inputs. The CutNoise makes a region of the short-exposure input image the same as ground truth; therefore, D2HNet learns to directly use the short-exposure input at this region. It makes D2HNet learn where to deblur and enhance long-exposure images in addition to how to deblur and enhance long-exposure images \cite{yoo2020rethinking}. Also, it helps balance the usage of long- and short-exposure inputs.

\section{More details of Noise Model}

\begin{figure*}[t]
\centering
\includegraphics[width=\linewidth]{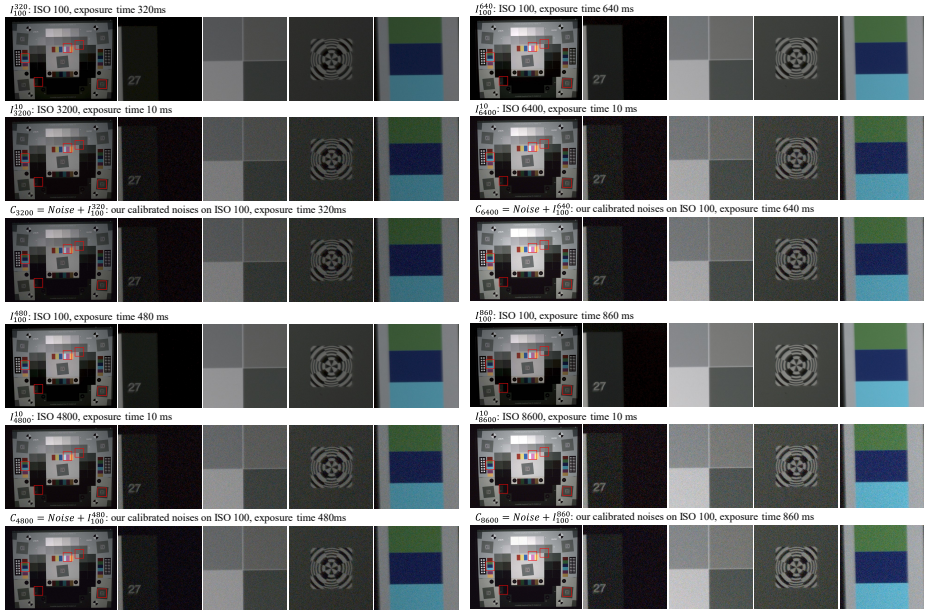}

\vspace{-0mm}

\caption{Illustration of noise calibration results on ISO 3200, 4800, 6400, and 12800. There are four patches selected from the greyworld chart for readers to compare specific regions: dark region, checkerboard edges, round edges, color blocks. $I_{100}^{*}$ denote photos captured under ISO 100, which we assume there are almost no noises. $I_{*}^{10}$ are photos with real noises. $C_*$ are the addition of calibrated noises on clean images with specific ISO values, i.e., $I_{100}^{*}$. The overall brightness is generally equal for $I_{100}^{*}$, $I_{*}^{10}$, and $C_*$ since ISO$\times$exposure time is equal. Please compare the patterns of real and calibrated noises.}
\label{noise}

\vspace{-0mm}

\end{figure*}

We use the physics-based noise model \cite{wei2020physics} to calibrate the \emph{Xiaomi Mi Note 10} smartphone for training the D2HNet. The ISO range of this smartphone is [100, 12800]. At the training, we randomly select the long-exposure ISO from [1000, 4000] and the short-exposure ISO from [6400, 12800] uniformly. It ensures that the noises in the long-exposure input are slighter than in the short-exposure input. At the validation, we add noises to clean validation images from D2-Dataset as inputs. The same ISO ranges are used for validation images. At the testing, since the D2HNet is trained with the calibrated noise model, it can directly enhance the long- and short-exposure image pair captured by the smartphone. We show some samples in Figure \ref{noise} to illustrate the noise calibration results.

\begin{figure*}[t]
\centering
\includegraphics[width=\linewidth]{pdf_supp_compressed/data.pdf}

\vspace{-0mm}

\caption{Illustration of the image synthesis pipeline of D2-Dataset and some examples.}
\label{data}

\vspace{-0mm}

\end{figure*}

\end{document}